\documentclass[conf]{new-aiaa}
\usepackage[utf8]{inputenc}
\usepackage[linesnumbered,ruled,vlined]{algorithm2e}
\usepackage{algpseudocode}
\usepackage{graphicx}
\usepackage{amsmath}
\usepackage[version=4]{mhchem}
\usepackage{siunitx}
\usepackage{longtable,tabularx}
\usepackage{float}
\usepackage{ragged2e}
\title{Kalman Filtering Based Flight Management System Modeling for AAM Aircraft}

\author{Balram Kandoria\footnote{Machine Learning Research Engineer, email: bkandoria@skygrid.com} and Aryaman Singh Samyal\footnote{Machine Learning Research Engineer, email: asamyal@skygrid.com}}
\affil{SkyGrid, Austin, TX, 78727}

\begin{document}

\maketitle

\begin{abstract}

Advanced Air Mobility (AAM) operations are planned to utilize strategic flight planning services that predict temporal uncertainties to validate flight plans against hazards such as weather cells, restricted airspaces, and CNS disruption areas. This paper presents a Kalman Filter-based uncertainty propagation method that models Flight Management System (FMS) correction behavior through a sigmoid-blended measurement noise covariance. The sigmoid formulation generalizes existing discrete FMS activation thresholds into a continuous, tunable function that smoothly transitions the filter's measurement noise based on progress toward each waypoint. When the measurement noise is high, due to an inverse relationship, the Kalman gain is small and thus uncertainty grows; as the aircraft nears a waypoint, measurement noise decreases as a function of progress, the Kalman gain increases, and state covariance contracts which models the FMS progressively correcting toward the planned trajectory. The approach is computationally efficient (up to two orders of magnitude faster than Monte Carlo methods), scales with control inputs, and is parametrically tunable for different classes of aircraft. The measurement noise covariance is calibrated using real Automatic Dependent Surveillance-Broadcast (ADS-B) data from commercial Instrument Flight Rules (IFR) flights serving as surrogates for future AAM operations, achieving coverage probability conservative relative to theoretical Gaussian predictions at the 1-$\sigma$ confidence level on a hold out verification dataset ($N = 36$). Parameter sensitivity analysis across multiple flight routes demonstrates robust behavior, and comparative evaluation against Monte Carlo and Linear Propagation methods contextualizes the method's computational and accuracy trade-offs.

\end{abstract}

% Comments
%

\section{Introduction}
\label{sec:intro}

Urban Air Mobility (UAM), a subset of Advanced Air Mobility (AAM) activities, is a new aviation sector with new vehicle entrants designed for short-duration flights within urban airspace and low-altitude flight paths not yet designated for IFR operations. These aircraft are Autonomous Flight Rules (AFR) enabled, as described in the Concept of Operations for Automated Flight Rules \cite{skygrid_afr_conops}, allowing them to self-separate, navigate, and manage contingencies with minimal human oversight. As a result, UAM aircraft are expected to proliferate quickly over the coming decades. The National Airspace System (NAS), which currently requires human oversight for each active flight, is not designed to handle the increased traffic density that UAM operations will introduce. UAM flights must be integrated into existing NAS infrastructure without overloading ATC operators while maintaining a level of safety consistent with standard ANSP practices. The FAA's UAM Concept of Operations \cite{faa_uam_conops_v2} outlines the operational framework for integrating these vehicles into controlled airspace, while the SkyGrid Concept of Operations \cite{skygrid_conops} details the service-oriented architecture required to support AFR-enabled flights, including strategic planning, conflict detection, and demand-capacity balancing.

With numerous UAM aircraft operating simultaneously, advanced planning and coordination services must be developed to ensure air traffic does not violate the capacity limits of shared resources or the separation minima with other operations. Strategic planning services play an integral role in qualifying submitted flight plans against the latest Notice to Airmen (NOTAMs), Airmens Meteorological Information (AIRMETs), Significant Meteorological Information (SIGMETs), and other weather hazards. Such services must predict how an aircraft will behave during a flight by understanding aircraft-specific dynamics, the flight environment, and the trajectory uncertainty induced from various internal and external sources. Traditional strategic planning algorithms typically consider a nominal operational environment and model disturbances using conservative linearized models, such as those of Dougui et al.\ \cite{LPA} and Banerjee and Corbetta \cite{nasa_uncertainty_quantification}, which tend to be overly conservative. Realistic and adaptive models for uncertainty are a necessity for strategic planning services. Accounting for the variance in the required time of arrival (RTA) at each waypoint provides planners the ability to better utilize the airspace and plan safe trajectories.

In this paper, an adaptation of existing uncertainty modeling concepts is proposed. The concept is applied to an uncertainty propagation engine, a linear Kalman Filter, to assigns uncertainty directly proportional to the size of control inputs into a system. Specifically, the traditional Kalman Filter is redefined to enable it to serve as a model for an advanced Flight Management System (FMS) associated with an UAM aircraft.

The primary contributions of this work are:
\begin{enumerate}
    \item A novel sigmoid-blended measurement noise covariance formulation that generalizes discrete FMS activation thresholds into a continuous, tunable function. This extends the Light Propagation Algorithm (LPA) concept of \cite{LPA} by replacing the discrete two-thirds activation threshold with a parametric sigmoid, enabling smooth transitions in FMS correction trust throughout the flight segment.
    \item A framework that repurposes the linear Kalman Filter as an FMS uncertainty model, where the measurement update step represents FMS corrections and the process noise represents unmodeled disturbances. The velocity covariance output is converted to temporal (RTA) variance following the method of \cite{nasa_uncertainty_quantification}.
    \item A data-driven calibration procedure using real ADS-B data from commercial IFR flights (as surrogates for UAM operations) to extract the measurement noise covariance matrix, enabling the model to be tuned for specific aircraft types or operational environments. Verification on a held-out dataset demonstrates conservative coverage probability relative to theoretical Gaussian bounds at the 1-$\sigma$ confidence level ($N = 36$).
\end{enumerate}

\section{Related Work}
\label{sec:related_work}

Aircraft state uncertainty estimation is a foundational research area for air traffic management (ATM) services. This section reviews prior work on spatial and temporal uncertainty modeling that informs the approach presented in this paper.

Paielli and Erzberger \cite{nasa_conflict_probability_estimation} established the framework for modeling aircraft position uncertainty in the context of conflict probability estimation. The authors decompose the position error into along-track (longitudinal) and cross-track (lateral) components. The along-track deviation is modeled with a linear growth model as a function of flight time, forming the basis for probabilistic separation assessment between aircraft pairs. The cross-track deviation is assigned a constant spatial error, justified by the effectiveness of modern lateral path-following control systems. This decomposition that separates longitudinal timing uncertainty from lateral tracking error remains a standard framework in subsequent work and motivates the temporal uncertainty focus of this paper.

Dougui et al. \cite{LPA} extend this framework by explicitly modeling the behavior of advanced FMS architectures. Their Light Propagation Algorithm (LPA) defines a velocity profile that diverges from nominal at a constant $\pm$ 1.06 growth rate until two-thirds of the required time of arrival (RTA) between waypoints has elapsed. Beyond this threshold, the model assumes that the FMS increases the corrective authority to bring the aircraft within a $\pm$ 10-second RTA tolerance. This discrete activation threshold captures the staged nature of traditional FMS correction logic.

The LPA method produces conservative uncertainty bounds, which is desirable for safety-critical planning applications. However, the constant growth rate cannot adapt to varying environmental conditions, e.g., gusty wind scenarios produce the same uncertainty growth as calm conditions. Additionally, the discrete two-thirds threshold produces sharp transitions that do not reflect the continuous correction behavior of modern FMS implementations.

Banerjee and Corbetta \cite{nasa_uncertainty_quantification} present methods for converting spatial (position and velocity) variances into temporal (RTA) variances. Their approach derives the time-of-arrival variance from the velocity covariance using a linear kinematic relationship. However, their method relies on empirical flight data from drone operations to directly obtain velocity variances. From a strategic planning perspective, such data are unavailable before flight execution, motivating the need for a predictive uncertainty propagation model, such as the Kalman Filter approach proposed in this work, that generates velocity covariance estimates from first principles.
\section{Methodology \& Mathematical Formulation}
\label{sec:methods}

\subsection{Flight Plan Fitting}

A continuous 4D trajectory representation is required to extract velocity and acceleration profiles between discrete waypoints. \cite{nasa_uncertainty_quantification} proposes non-uniform rational B-splines (NURBS) for this purpose. However, NURBS does not guarantee monotonicity in the time domain when fitting 4D waypoints, which can produce infeasible trajectories where time decreases between spatial positions.

\begin{figure}[H]
    \centering
    \includegraphics[width=\textwidth]{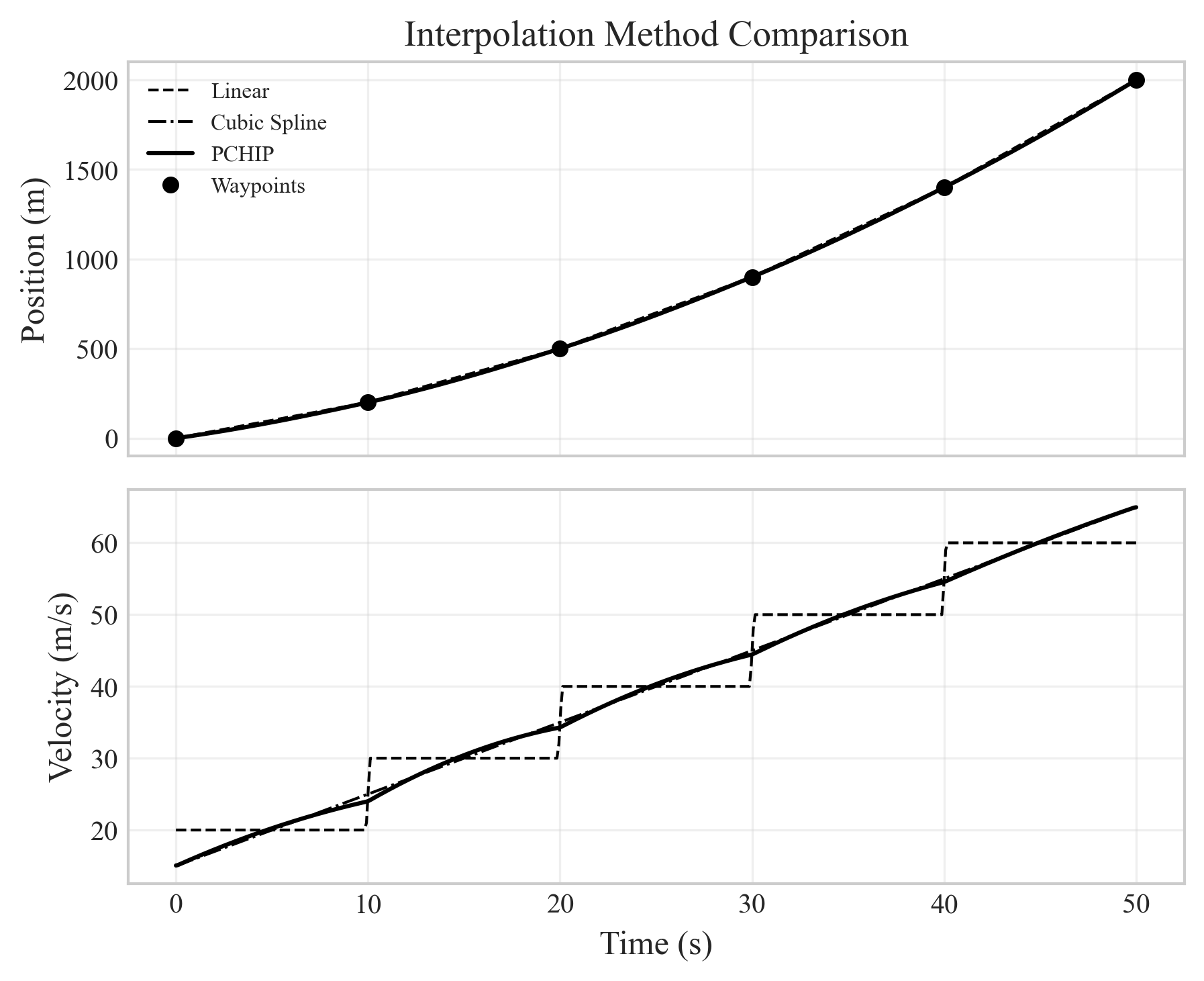}
    \caption{Comparison of interpolation methods showing position (top) and velocity (bottom) profiles. Linear interpolation produces velocity discontinuities; cubic splines and PCHIP produce smooth profiles, with PCHIP additionally preserving monotonicity.}
    \label{fig:interpolation_comparison}
\end{figure}

A Piecewise Cubic Hermite Interpolating Polynomial (PCHIP) \cite{fritsch1980monotone} is adopted as the interpolation method. PCHIP was selected based on three criteria: (1) it strictly enforces monotonicity, guaranteeing that the time parameterization remains physically valid; (2) it is shape-preserving, avoiding the oscillatory behavior (Runge's phenomenon) that cubic splines exhibit near sharp changes in the data; and (3) it provides $C^1$ continuity, yielding smooth velocity profiles suitable for the Kalman Filter's control input requirements. Figure~\ref{fig:interpolation_comparison} compares the position, velocity, and acceleration profiles produced by each method on an example waypoint set.

The velocity and acceleration values, extracted via differentiation, serve as the control input for the Kalman Filter. The planned position at each time step is used as the measurement in the update step. Although this position is not a true sensor measurement, it represents the FMS target that the aircraft is tracking, and thus models the correction behavior of the FMS toward the planned trajectory.

\subsection{Variance Extraction}

Following \cite{nasa_uncertainty_quantification} without modification, a linear kinematic model is used to describe vehicle motion between waypoints and convert velocity variance to temporal variance:
\begin{align*}
\Delta p_k \approx v_k \Delta t_{a,k}
\end{align*}
\begin{align*}
\Delta p_k &= p_k - p_{k-1}
&&\text{: vector distance between waypoint } k \text{ and } k-1, \\
\Delta t_{a,k} &= t_{a,k} - t_{a,k-1}
&&\text{: difference between the RTA for the pair of waypoints}, \\
v_k &= [v_{x,k}, v_{y,k}, v_{z,k}]
&&\text{: cruise speed vector between waypoint } k \text{ and } k-1.
\end{align*}

The differential RTA at each waypoint is calculated by inverting the above equation and computing the norm of the velocity vector:
\begin{align*}
\Delta t_{a,k} \approx \frac{\rVert \Delta p_k \rVert_2}{\rVert v_k \rVert_2}
\end{align*}

The variance of the time of arrival at each waypoint $k$ is computed by cumulative sum of the covariances up to waypoint $k$:
\begin{align*}
\sigma^2_{a,k} = \begin{cases}
      \Delta t_{a,k} \frac{\rVert\sigma^2_{v_k}\rVert_2}{\rVert v_k \rVert_2 }\quad \forall \qquad \sigma^2_{v_k},v_k : \rVert \sigma^2_{v_k} \rVert_2 / \rVert v_k \rVert_2 < \delta \bar{v_0}\\
      0 \qquad \text{otherwise}
   \end{cases}
\end{align*}
\begin{align*}
\sigma^2_{t_{a,k}} = \sum^k_{i=1}\sigma_{i,j}
\end{align*}

The criterion shown above represents the minimum noise threshold allowed to influence the time variance. With the variance in time described, the RTA is modeled as a normal probability density function:
\begin{align*}
t_{a,k} \sim \mathcal{N}(\bar{t_{a,k}},\sigma^2_{t_{a,k}})
\end{align*}

This enables the generation of upper and lower RTA estimates at each waypoint based on a desired confidence level.

\subsection{Uncertainty Modeling}

The velocity variance is generated using a standard linear Kalman Filter as the uncertainty propagation engine. A Kalman Filter is used not as a state estimator in the traditional sense, but as a model of the FMS correction process: the predict step models uncorrected trajectory drift, and the update step models FMS corrections toward planned waypoints. The filter design is as follows:

\textbf{State Vector:}
\begin{equation*}
\mu = \begin{bmatrix} \mathbf{p} \\ \mathbf{v} \end{bmatrix} = \begin{bmatrix} x & y & z & v_x & v_y & v_z \end{bmatrix}^T
\end{equation*}

\textbf{System Matrices} (expressed in $3\times3$ block form with $I_3$ the identity and $0_3$ the zero matrix):
\begin{equation*}
A = \begin{bmatrix} I_3 & \Delta t\, I_3 \\ 0_3 & I_3 \end{bmatrix}, \quad
B = \begin{bmatrix} \Delta t\, I_3 \\ 0_3 \end{bmatrix}, \quad
C = \begin{bmatrix} I_3 & 0_3 \end{bmatrix}
\end{equation*}

\textbf{Process Noise Covariance} (piecewise-constant white-noise acceleration model):
\begin{equation*}
R = \sigma_a^2 \begin{bmatrix} \tfrac{1}{4}\Delta t^4\, I_3 & \tfrac{1}{2}\Delta t^3\, I_3 \\ \tfrac{1}{2}\Delta t^3\, I_3 & \Delta t^2\, I_3 \end{bmatrix}
\end{equation*}

The process noise covariance matrix models white-noise acceleration to capture unmodeled forces. The variable $\sigma_a^2$ is the acceleration noise variance; a larger value represents higher uncertainty, such as windy conditions or strong external disturbances.

The state space described above represents both position and velocity in 3D space. The control inputs are velocity and acceleration; however, acceleration has no effect on the filter's internal state and is not considered in this implementation.

\subsection{Sigmoid-Blended Measurement Noise}

\begin{figure}[H]
    \centering
    \includegraphics[width=\textwidth]{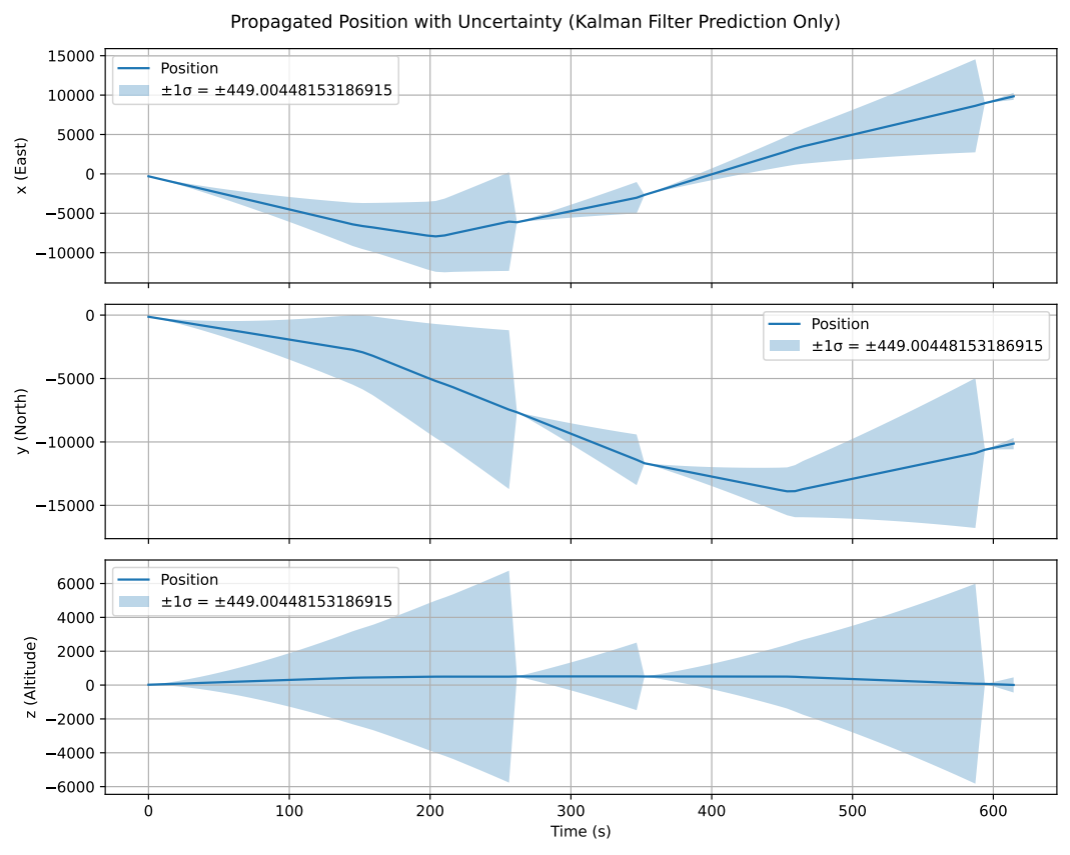}
    \caption{Kalman filter uncertainty propagation with discrete LPA threshold at $p = 2/3$, showing an abrupt drop in position uncertainty when the update step activates.}
    \label{fig:kf_steep_drop}
\end{figure}

Extending the uncertainty Light Propagation Algorithm (uLPA) of \cite{LPA}, a continuous sigmoid-blended measurement noise covariance, which generalizes the discrete FMS activation threshold, is proposed. In \cite{LPA}, FMS corrections activate at two-thirds of the progress between waypoints. The concept is implemented within the Kalman Filter framework: before FMS activation, only the predict step executes and covariance grows unbounded; after activation, the update step engages and covariance contracts. However, applying the discrete threshold directly produces an unrealistic discontinuity in the uncertainty profile (Fig.~\ref{fig:kf_steep_drop}).

An alternative approach is needed to eliminate this discontinuity. A sigmoid-blended measurement noise covariance, the primary novel contribution of this work, can be utilized as a new model representing FMS correction authority. The measurement noise covariance $Q$ is bounded between $Q_{min}$ (full FMS correction, low noise) and $Q_{max}$ (no FMS correction, high noise):
\[
Q_{min} = I_3 \cdot 10^0, \quad Q_{max} = I_3 \cdot 10^8
\]
where $Q_{max}$ represents the measurement uncertainty in the absence of FMS corrections and $Q_{min}$ represents the residual uncertainty under full FMS control. The value of $Q_{max}$ is tunable and can be calibrated from flight data for specific aircraft types or environmental conditions (see Section~\ref{sec:validation}).

A sigmoid function smoothly blends between these bounds as a function of progress ratio $p \in [0,1]$ within each waypoint segment:
\begin{equation}
    \sigma(p) = \frac{1}{1+e^{k(p - p_{LPA})}}
    \label{eq:sigmoid}
\end{equation}
\begin{equation}
    Q(p) = Q_{min} + (Q_{max} - Q_{min})\cdot \sigma(p)
    \label{eq:qp}
\end{equation}
where $k$ controls the steepness of the transition and $p_{LPA}$ is the activation threshold. This formulation generalizes the discrete LPA approach: setting $p_{LPA} = 2/3$ with large $k$ recovers the original discrete behavior, while $p_{LPA} = 0$ yields continuous FMS correction from the start of each segment. The parameter $p_{LPA}$ thus controls where along the segment the FMS begins actively correcting, and $k$ controls how gradual that transition is.

The physical reasoning is as follows: when $Q$ is large (far from waypoint), the Kalman gain $K = \Sigma C^T(C\Sigma C^T + Q)^{-1}$ is small, so the update step has minimal effect on the state covariance, uncertainty grows. As the aircraft approaches the waypoint, $Q$ decreases while $K$ increases, and the update step progressively reduces the state covariance. This models the FMS by gradually increasing its corrective authority as the aircraft approaches its target.

\begin{figure}[H]
    \centering
    \includegraphics[width=\textwidth]{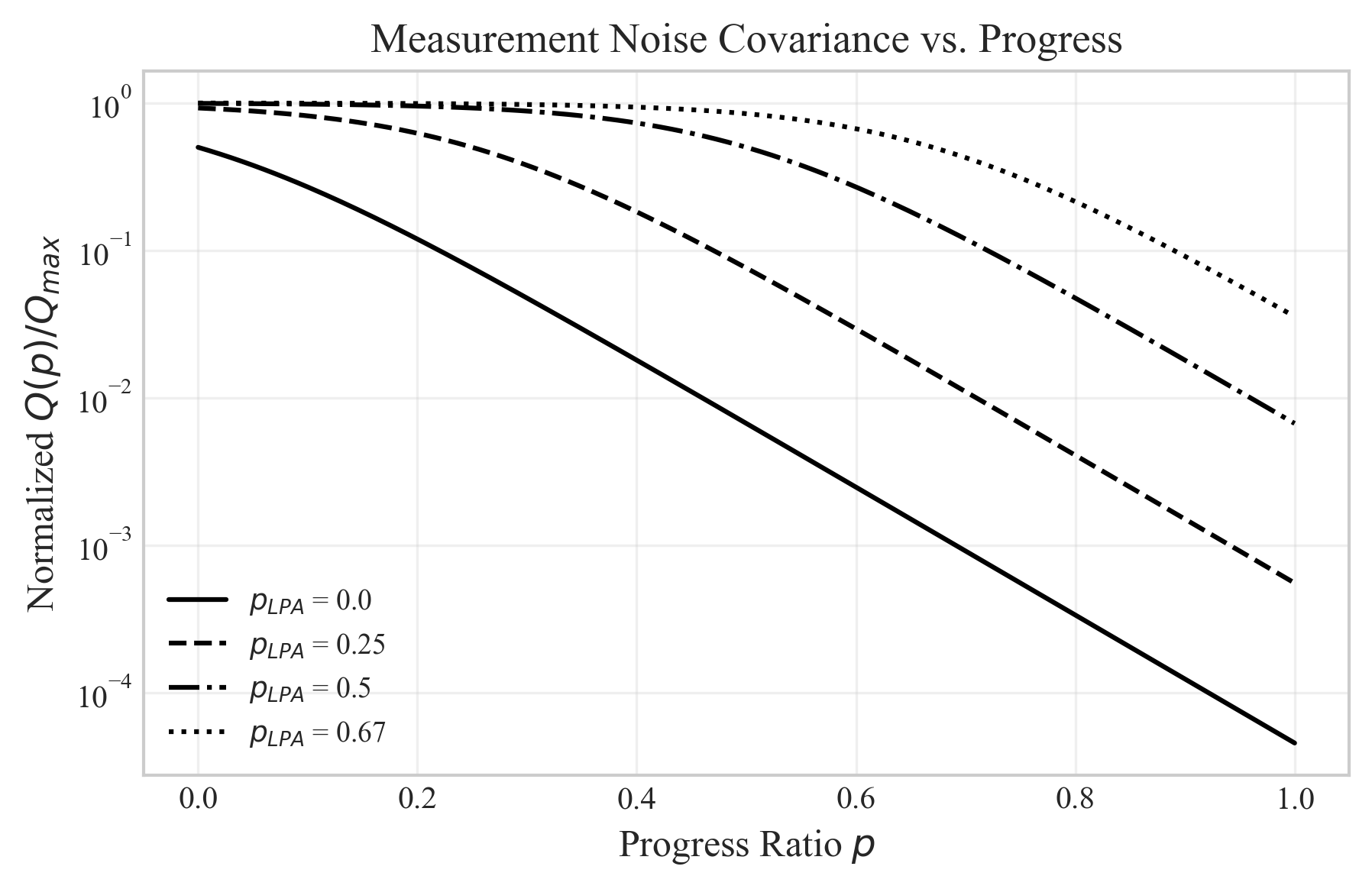}
    \caption{Normalized measurement noise covariance $Q(p)/Q_{max}$ for different LPA activation thresholds. Setting $p_{LPA} = 0$ yields continuous correction; $p_{LPA} = 0.67$ approximates the discrete LPA behavior.}
    \label{fig:sigmoid_parametric}
\end{figure}

Figure~\ref{fig:sigmoid_parametric} shows the normalized $Q(p)$ for several values of $p_{LPA}$, demonstrating the generalization. Modern FMS systems correct deviations continuously rather than activating at a fixed threshold; accordingly, we adopt $p_{LPA} = 0$ as the preferred configuration, which provides the smoothest uncertainty evolution.

\begin{figure}[H]
    \centering
    \includegraphics[width=\textwidth]{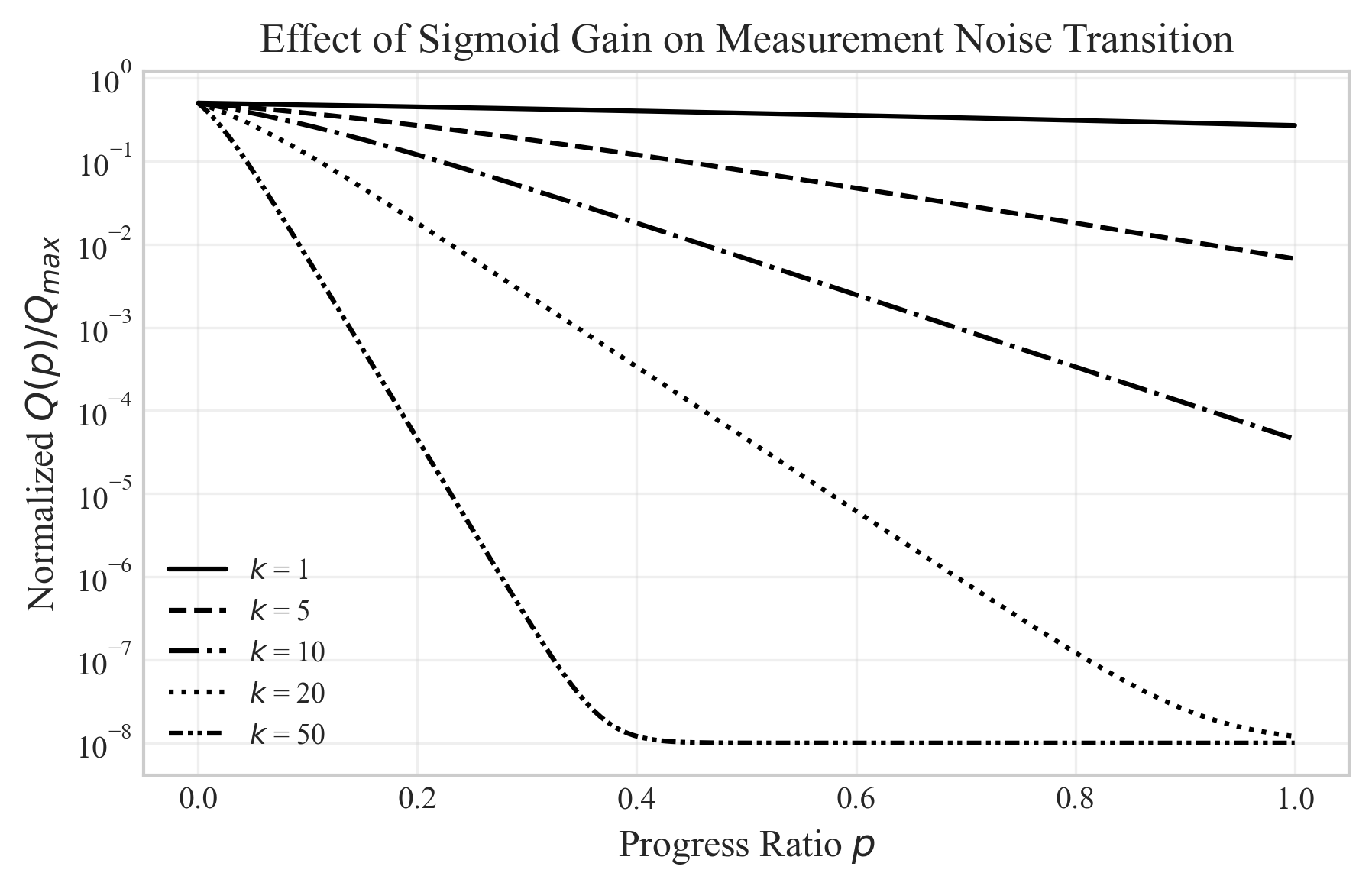}
    \caption{Effect of sigmoid gain $k$ on the measurement noise transition. Higher $k$ produces a sharper transition between $Q_{max}$ and $Q_{min}$.}
    \label{fig:sigmoid_gain}
\end{figure}

The effect of the sigmoid gain parameter $k$ is shown in Fig.~\ref{fig:sigmoid_gain}. Small values of $k$ produce a gradual transition across the full segment; large values produce a sharper switch. The value $k = 10$ provides a moderate transition rate, where FMS control authority is meaningful but won't reduce uncertainty drastically, which is used throughout this work.

\begin{figure}[H]
    \centering
    \includegraphics[width=\textwidth]{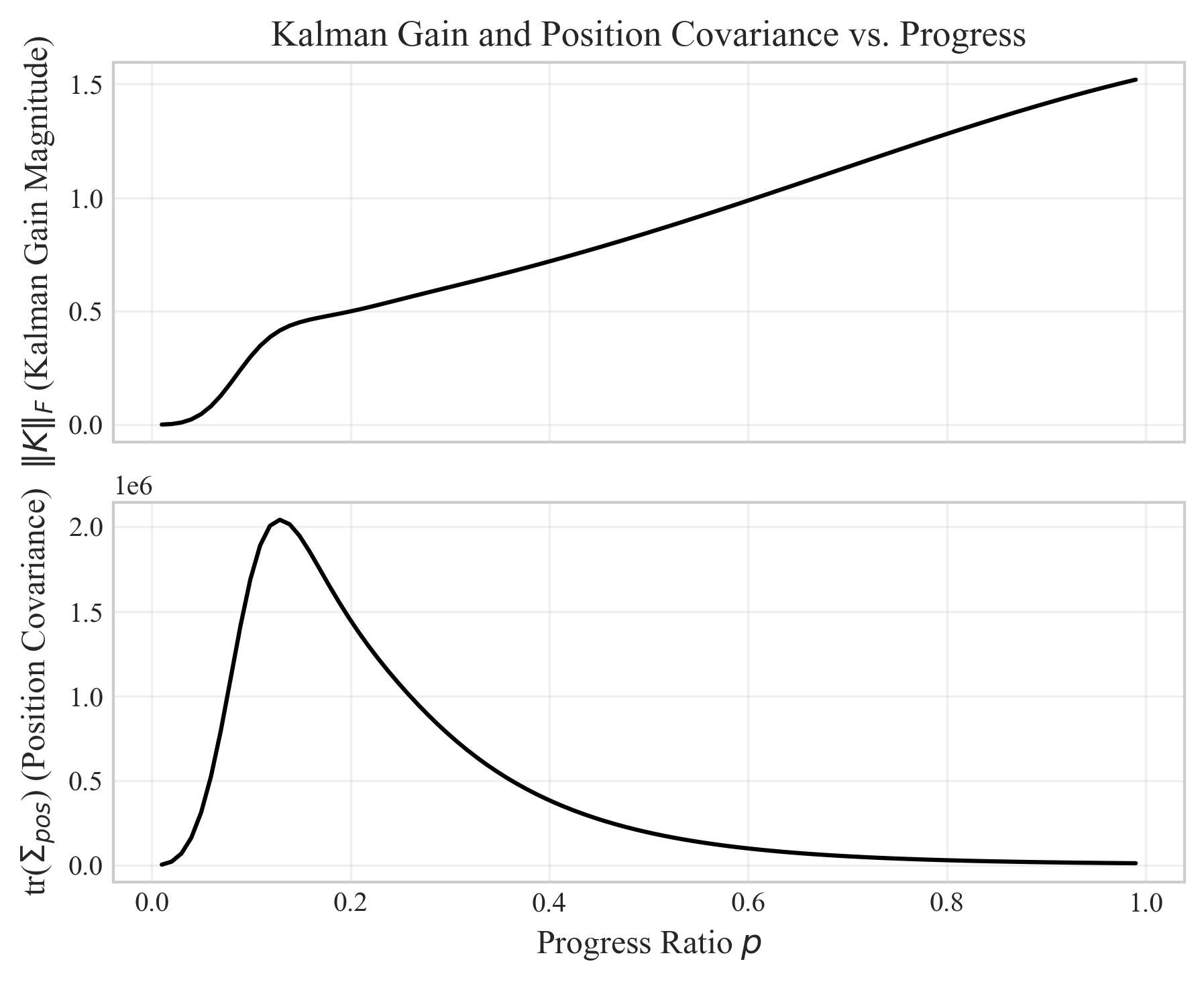}
    \caption{Kalman gain magnitude (top) and position covariance trace (bottom) vs.\ progress ratio for a single segment with $p_{LPA} = 0$, $k = 10$.}
    \label{fig:kalman_gain_progress}
\end{figure}

Figure~\ref{fig:kalman_gain_progress} illustrates the resulting Kalman gain magnitude and position covariance trace as functions of progress ratio, showing how the gain increases (and covariance decreases) as the aircraft approaches each waypoint.

\begin{figure}[H]
    \centering
    \includegraphics[width=\textwidth]{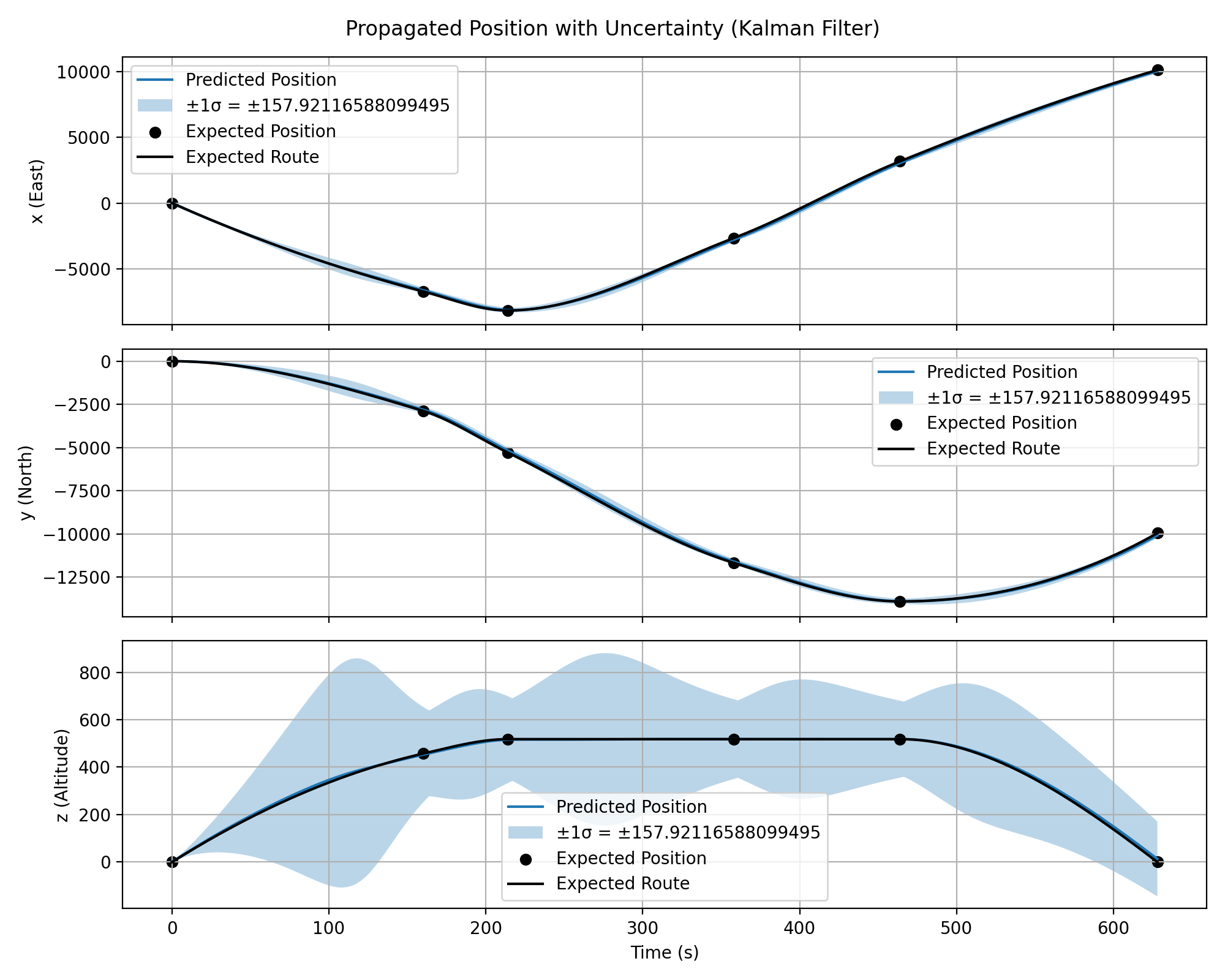}
    \caption{Kalman filter uncertainty propagation using sigmoid-blended measurement noise with $p_{LPA} = 0$, showing smooth and continuous uncertainty evolution.}
    \label{fig:kf_propagation}
\end{figure}

With $p_{LPA} = 0$, the resulting uncertainty propagation exhibits smooth, continuous evolution as shown in Fig.~\ref{fig:kf_propagation}. The state covariance grows during early portions of each segment and contracts as the FMS correction strengthens near the waypoint.

\subsection{RTA Variance and Covariance Decomposition}

\begin{figure}[H]
    \centering
    \includegraphics[width=\textwidth]{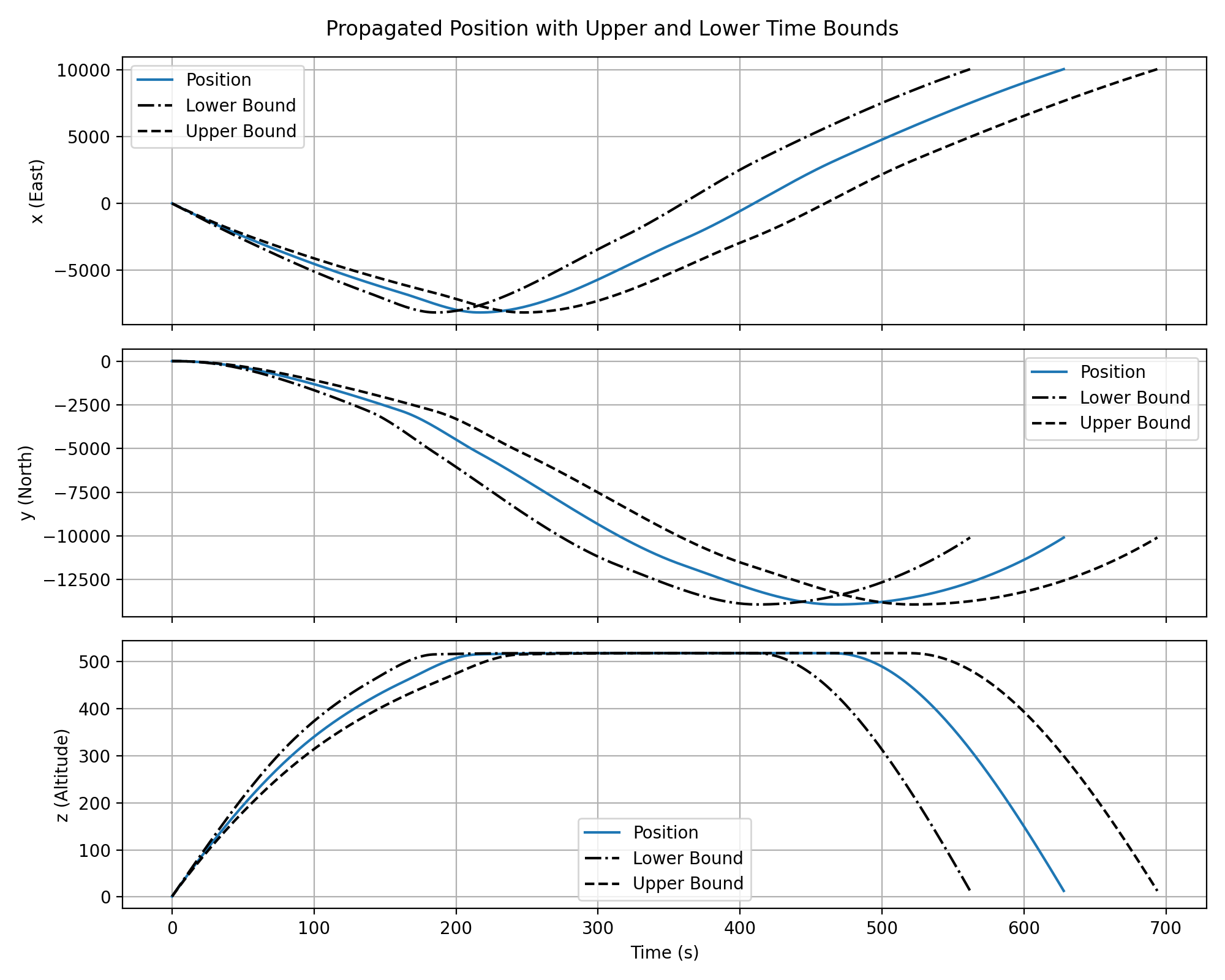}
    \caption{Required time of arrival (RTA) predictions showing nominal trajectory with upper and lower temporal variance bounds derived from velocity covariance propagation.}
    \label{fig:rta_bounds}
\end{figure}

Following \cite{nasa_uncertainty_quantification}, the velocity covariance from the Kalman Filter is converted to temporal (RTA) variance. Figure~\ref{fig:rta_bounds} shows the resulting upper and lower RTA bounds for an example route.

\begin{figure}[H]
    \centering
    \includegraphics[width=\textwidth]{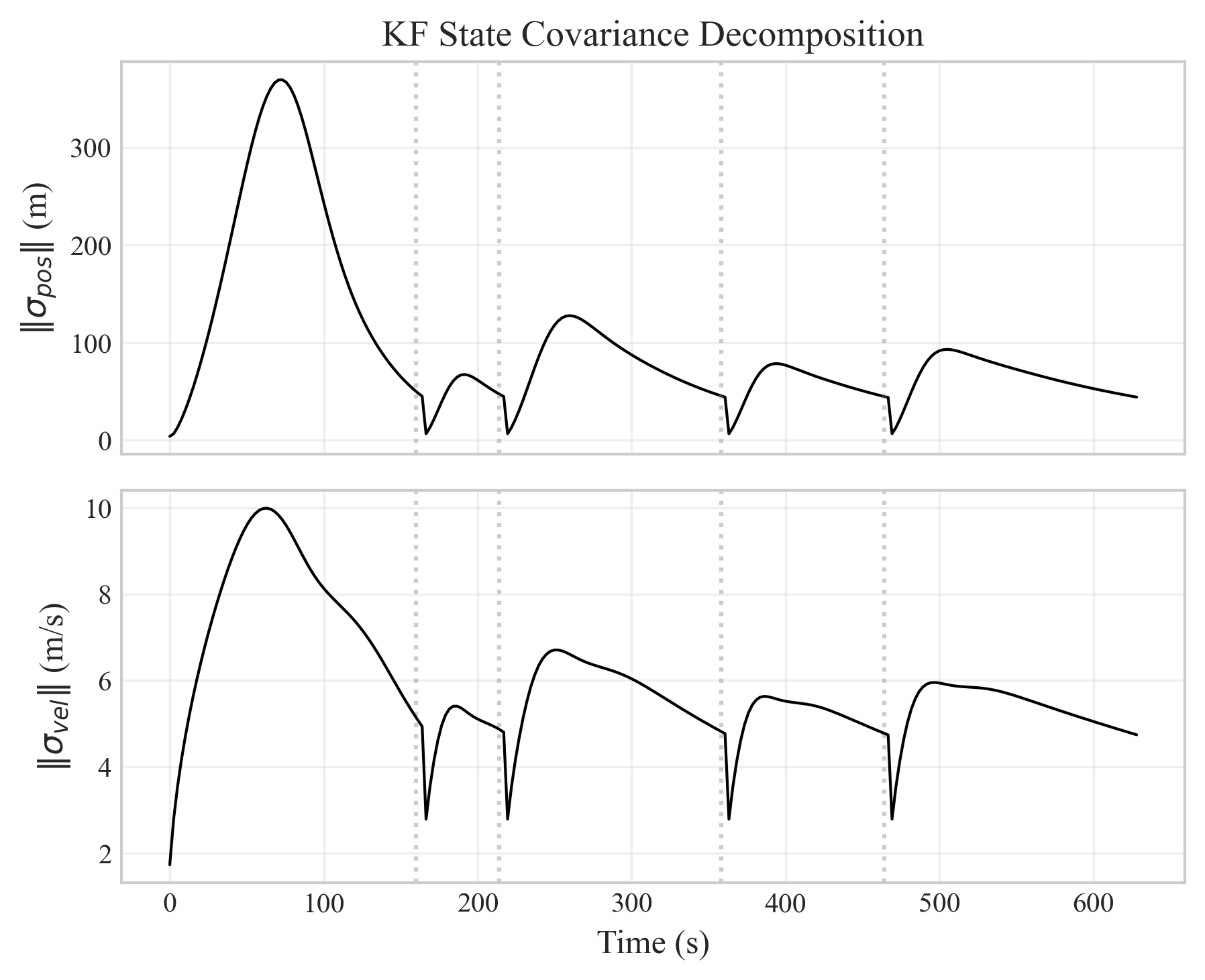}
    \caption{Decomposition of KF state covariance behavior. Top: position standard deviation showing growth/contraction cycles. Bottom: velocity standard deviation. Vertical dashed lines indicate waypoint positions.}
    \label{fig:variance_decomposition}
\end{figure}

It is important to distinguish two quantities: the KF state covariance and the cumulative RTA variance. The state covariance (position and velocity) exhibits growth during the predict phase and contraction during the update phase within each waypoint segment. The cumulative RTA variance, however, is computed as a running sum of non-negative time increments and is therefore monotonically increasing by construction. The contraction in state covariance manifests as a reduced \emph{growth rate} of the RTA variance, not as an absolute decrease. Figure~\ref{fig:variance_decomposition} decomposes these quantities, showing the position standard deviation, velocity standard deviation, and the RTA variance growth rate over time. The growth rate clearly decreases near waypoints where the update step is most active.

\subsection{Measurement Noise Calibration}
\label{sec:validation}

The measurement noise covariance $Q_{max}$ must be calibrated from real flight data to represent realistic FMS tracking behavior. In the absence of publicly available UAM flight data, we use commercial IFR flights operated by Boeing 737-series and Airbus A3xx-series aircraft as surrogates. These aircraft employ automated FMS systems that actively track filed flight plans with high fidelity; the same behavior that UAM vehicles will exhibit with similar or more advanced avionics. VFR flights are explicitly excluded because they do not reliably follow filed paths, which would contaminate the covariance estimate with deviations unrelated to FMS tracking performance.

The calibration pipeline uses two data sources: FAA-filed IFR flight plans retrieved from the System Wide Information Management (SWIM) feed via a Hive data warehouse, and corresponding ADS-B surveillance tracks retrieved from FlightRadar24. For each flight, both sources are matched by callsign and departure date, yielding paired planned-vs-actual trajectory data.

The covariance extraction procedure operates as follows. A set of $K = 100$ flight plans is queried from SWIM for the calibration date (June 20, 2025). For each qualifying flight $k$, the corresponding ADS-B track is retrieved, and both trajectories are time-aligned using Piecewise Cubic Hermite Interpolating Polynomial (PCHIP) interpolation onto a common fine-resolution time grid spanning their temporal overlap. Both interpolated paths are then converted from geodetic coordinates to an East-North-Up (ENU) local tangent plane frame. The position error vector is computed as $\mathbf{e}_k(t) = \mathbf{p}^{FAA}_k(t) - \mathbf{p}^{ADS\text{-}B}_k(t)$, and the per-flight empirical covariance is $\Sigma_k = \text{Cov}(\mathbf{e}_k)$. Flights with mean position error exceeding 5000 m are discarded as outliers (e.g., re-routed or diverted flights). The calibrated measurement noise is then the ensemble average:
\begin{equation}
    Q_{max} = \frac{1}{K}\sum_{k=1}^{K}\Sigma_k
    \label{eq:qmax_calibration}
\end{equation}
Because ADS-B altitude data has insufficient fidelity for independent calibration at the resolution required, the vertical variance is set to the mean of the horizontal components: $Q_{max}[2,2] = (Q_{max}[0,0] + Q_{max}[1,1])/2$. Of the 100 queried flight plans, $K = 12$ were IFR flights with less than 5000 m of path deviation, yielding the calibrated result:
\begin{equation}
    Q_{max} = \begin{bmatrix} 1.04 \times 10^8 & 8.63 \times 10^7 & 0 \\ 8.63 \times 10^7 & 1.17 \times 10^8 & 0 \\ 0 & 0 & 1.11 \times 10^8 \end{bmatrix} \text{ m}^2
    \label{eq:qmax_result}
\end{equation}
The diagonal elements correspond to position standard deviations of approximately 10.2 km (East) and 10.8 km (North), with a cross-correlation coefficient of 0.78. These magnitudes reflect the cumulative deviation between filed and actual trajectories over full flight durations (1--10 hours), not instantaneous tracking error.

The calibration and verification datasets are temporally separated, thereby preventing overfitting and ensuring dataset independence. The calibration uses flights from June 20, 2025 ($K = 100$ flight plans queried, 12 qualifying), while verification (Section~\ref{sec:test_setup}) uses a different date, June 22, 2025 (150 flight plans queried, 36 qualifying). 

A variance scaling parameter $\gamma \leq 1$ is applied to the cumulative RTA variance to account for the systematic overestimation inherent in summing 3D velocity contributions for primarily along-track motion. Crucially, $\gamma$ adjusts only the \emph{scale} of the uncertainty; the \emph{shape}, the segment-wise growth-and-contraction profile driven by the sigmoid-blended KF, is determined entirely by the physics-based model and is not tunable. Any uncertainty propagation method can be scaled post-hoc to hit a coverage target; the contribution here is a physically motivated temporal structure that $\gamma$ merely calibrates to a desired conservatism level. The coverage-vs-$\gamma$ relationship (Section~\ref{sec:results}) allows operators to select the desired conservatism level for their application.

\subsection{Methodology Summary}

The complete processing pipeline is summarized as follows:
\begin{enumerate}
    \item \textbf{Path Interpolation} (standard technique): Discrete waypoints are interpolated using PCHIP to produce a continuous 4D trajectory. Velocity and acceleration profiles are extracted via differentiation.
    \item \textbf{KF Propagation with Sigmoid-Blended $Q(p)$} (this work): A linear Kalman Filter propagates position and velocity uncertainty along the trajectory. The measurement noise covariance transitions from $Q_{max}$ to $Q_{min}$ via Eq.~\ref{eq:qp}, modeling progressive FMS correction.
    \item \textbf{Velocity-to-Time Variance Conversion} (adapted from \cite{nasa_uncertainty_quantification}): The velocity covariance from the KF output is converted to temporal variance using a linear kinematic relationship.
    \item \textbf{Variance Scaling} (calibrated): The cumulative RTA variance is scaled by $\gamma$ to match the desired coverage level, as determined during calibration against real flight data.
\end{enumerate}

\begin{figure}[H]
    \centering
    \includegraphics[width=\textwidth]{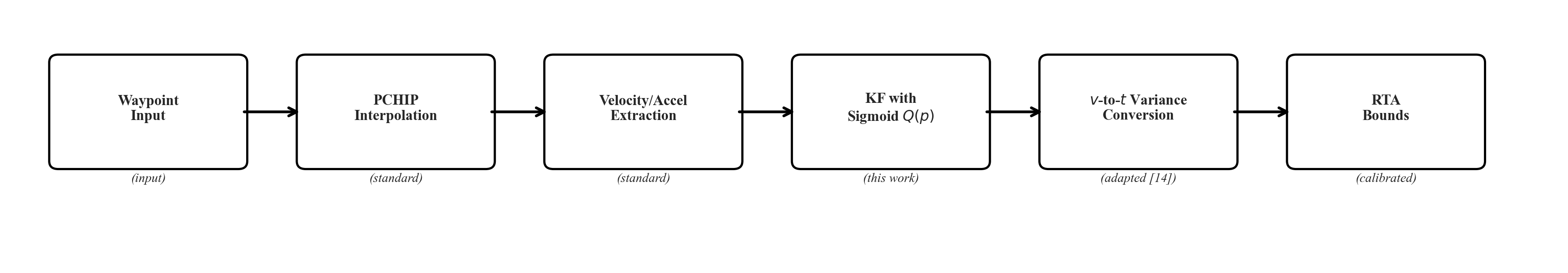}
    \caption{Processing pipeline with provenance. Each stage is labeled with its source: novel contributions (``this work''), adapted methods (with citation), or standard techniques.}
    \label{fig:block_diagram}
\end{figure}

Figure~\ref{fig:block_diagram} provides a visual summary of this pipeline with provenance annotations.

\section{Results}
\label{sec:results}

\subsection{Parameter Sensitivity Analysis}

The KF-FMS method is parameterized primarily via three parameters: acceleration noise $\sigma_a^2$, variance scaling parameter $\gamma$, and LPA activation threshold $p_{LPA}$. A sensitivity analysis was performed to determine the influence and tunability of the KF-FMS response to the value of individual parameters. In this sensitivity study, each parameter was independently swept while the remaining parameters were kept at their nominal values ($\sigma_a^2 = 0.5$, $\gamma = 0.7$, $p_{LPA} = 0$).

\begin{figure}[H]
    \centering
    \includegraphics[width=\textwidth]{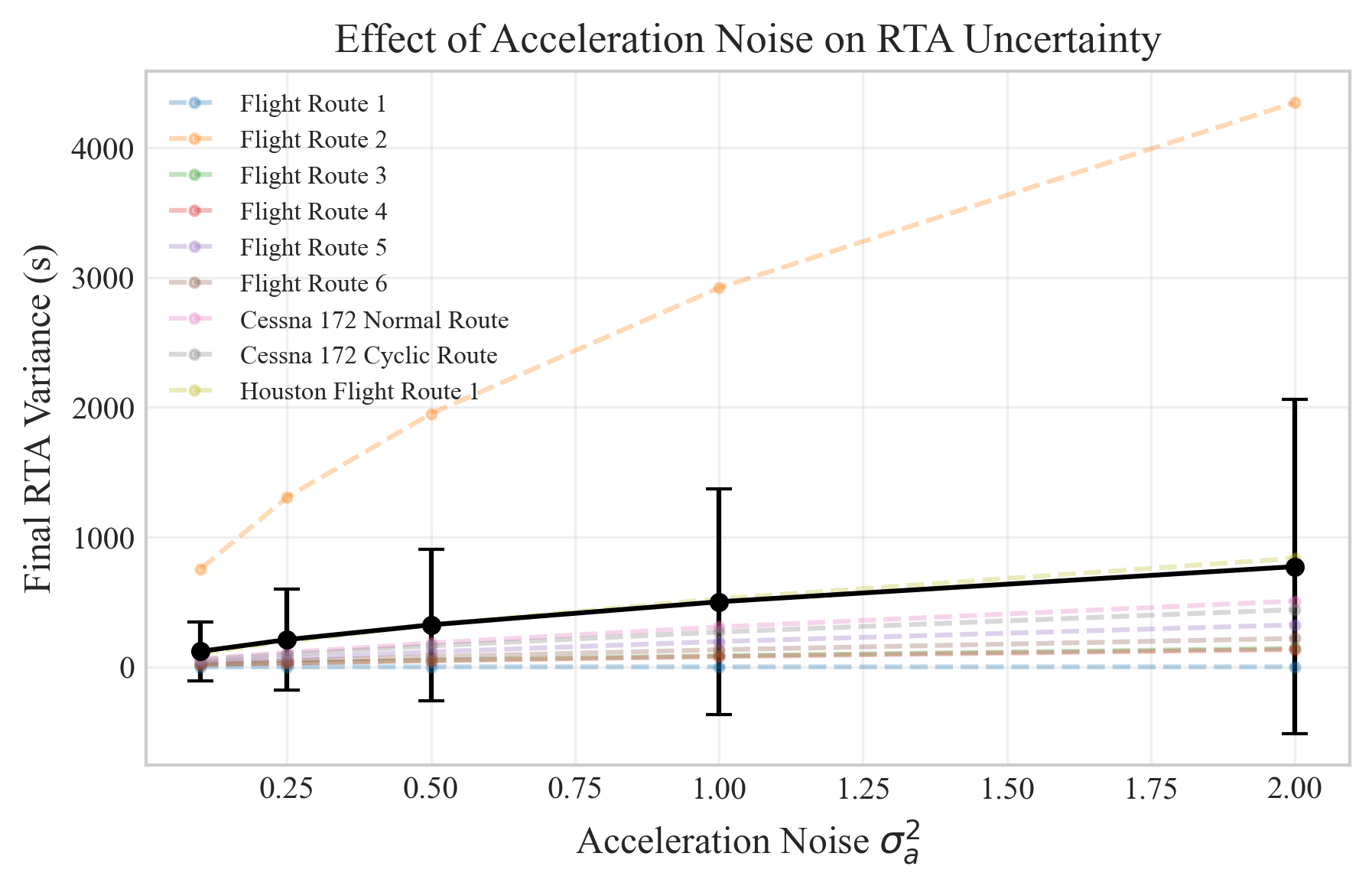}
    \caption{Effect of acceleration noise $\sigma_a^2$ on final RTA variance.}
    \label{fig:sensitivity_accel}
\end{figure}

Figure~\ref{fig:sensitivity_accel} shows the effect of varying acceleration noise on RTA uncertainty. The black line represents the mean across all routes, error bars indicate $\pm 1$ standard deviation, and dashed lines show individual routes. Extrapolated from the figure, an approximately linear relationship exists between acceleration noise and the final RTA variance; thus process noise directly scales the velocity covariance growth rate.

\begin{figure}[H]
    \centering
    \includegraphics[width=\textwidth]{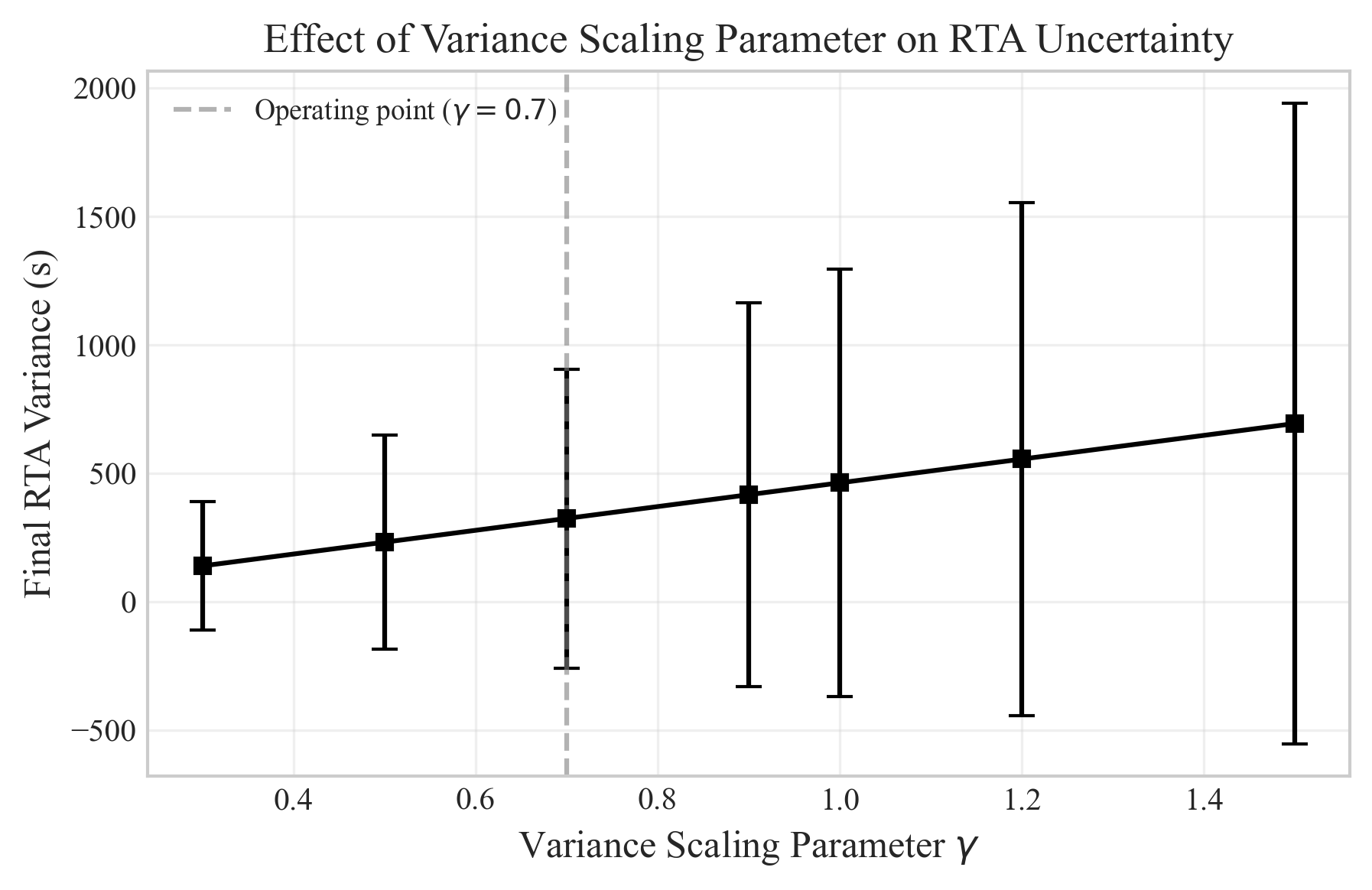}
    \caption{Effect of variance scaling parameter $\gamma$ on final RTA variance.}
    \label{fig:sensitivity_gamma}
\end{figure}

Figure~\ref{fig:sensitivity_gamma} shows the effect of the variance scaling parameter $\gamma$. As expected, the relationship is directly proportional since $\gamma$ is a multiplicative factor applied after calculating the cumulative variance. The operating point $\gamma = 0.7$ is marked, corresponding to the set value calibrated against ADS-B validation data.

\begin{figure}[H]
    \centering
    \includegraphics[width=\textwidth]{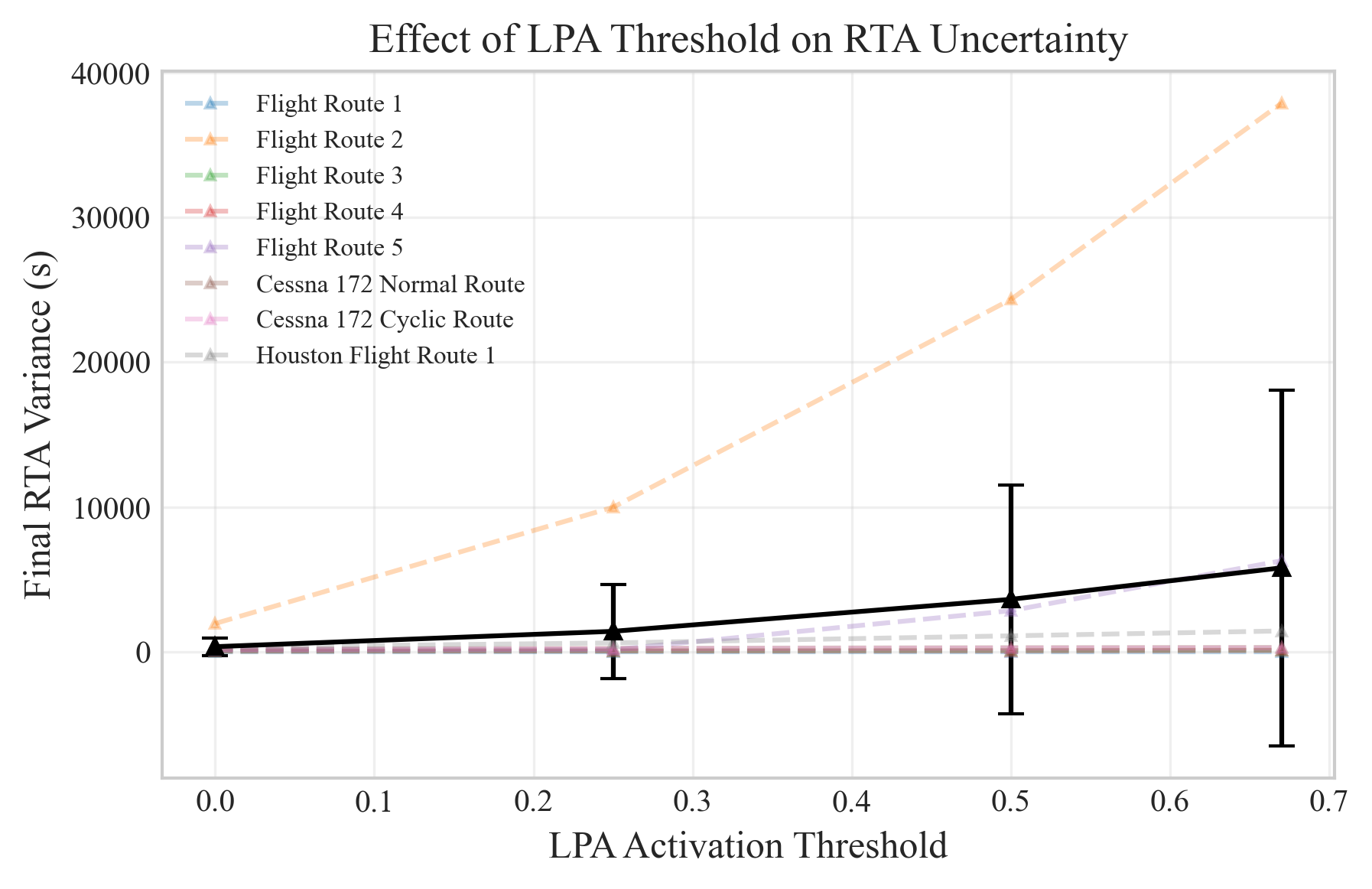}
    \caption{Effect of LPA activation threshold $p_{LPA}$ on final RTA variance.}
    \label{fig:sensitivity_lpa}
\end{figure}

Figure~\ref{fig:sensitivity_lpa} shows the effect of the LPA activation threshold. Higher thresholds delay the onset of FMS correction, resulting in greater accumulated uncertainty. The difference between $p_{LPA} = 0$ and $p_{LPA} = 0.67$ demonstrates the effect of the generalization: continuous correction produces approximately 30--50\% lower final uncertainty compared to the discrete two-thirds threshold. The figure also demonstrates that as the $p_{LPA}$ value increases, the value of the RTA variance becomes overly conservative and thus does not accurately represent true variance. 

\subsection{Verification Test Setup}
\label{sec:test_setup}

The $Q_{max}$ outputted from the KF tuning process, as described in Section~\ref{sec:validation}, was derived from a set of real-world ADS-B flight data recorded on June 20, 2025. The verification phase, which tests the predictive accuracy of the tuned $Q_{max}$ matrix, uses ADS-B flight data recorded on June 22, 2025 to ensure dataset independence. \\

As part of the required data buffering, the elapsed-time waypoints of each flight plan were converted to a common time axis starting from departure as described in Section~\ref{sec:validation}.\\

Flights are included in the verification set only if they satisfy at least one of: (1) 2D position RMSE between planned and actual paths $< 2000$ m, or (2) absolute arrival time difference $< 1000$ s. This is stricter than the calibration filter (mean error $< 5000$ m), ensuring that only flights with reliable trajectory data are used to evaluate predictive accuracy. From 150 queried flight plans (June 22, 2025), 36 flights passed the quality filter and had valid matched ADS-B data, forming the verification dataset. These flights span a range of domestic US routes with flight durations from approximately 60 to 590 minutes. \\

For each qualifying flight, the KF-FMS method propagates uncertainty along the filed flight plan using the calibrated $Q_{max}$ to produce a predicted RTA standard deviation $\sigma_t$ at the destination. A flight ``passes'' the verification if the actual ADS-B arrival time falls within $\pm \gamma \sigma_t$ of the FAA-planned arrival time. The pass rate across all qualifying flights quantifies the model's predictive reliability at the 1-$\sigma$ level.\\

\subsection{Empirical Verification Results}

Using the test setup described above, the KF-FMS method was evaluated against $N = 36$ held-out flights. For each flight, the predicted RTA variance bound was compared against the actual arrival time deviation observed via ADS-B.

Table~\ref{tab:verification_results} presents the per-flight verification results. For each flight, the signed arrival time deviation $\Delta t = t_{ADS\text{-}B} - t_{FAA}$ is compared against the predicted 1-$\sigma$ bound $\gamma\sigma_t$. The margin column indicates surplus capacity (positive) or overshoot (negative). Flights with negative margin exceed the predicted bound, corresponding to the 22.2\% of flights outside the 1-$\sigma$ interval. The failed flights exhibit large deviations (400--833 s) relative to short flight durations, consistent with significant rerouting or ATC delays not captured by the filed flight plan.

\begin{table}[H]
\centering
\caption{Per-Flight Verification Results ($\gamma = 0.7$, June 22, 2025)}
\label{tab:verification_results}
\footnotesize
\begin{tabular}{|l|r|r|r|r|c|}
\hline
Callsign & Duration (min) & $\Delta t$ (s) & $\gamma\sigma_t$ (s) & Margin (s) & Pass \\
\hline
ASA459 & 314 & +755 & 652 & $-$103 & $\times$ \\
ASA299 & 298 & +424 & 588 & +164 & $\checkmark$ \\
BOX383 & 593 & $-$90 & 2593 & +2503 & $\checkmark$ \\
AAL3074 & 108 & $-$647 & 271 & $-$376 & $\times$ \\
DAL2838 & 78 & +40 & 178 & +138 & $\checkmark$ \\
DAL2302 & 63 & +29 & 158 & +129 & $\checkmark$ \\
JBU1015 & 194 & +327 & 499 & +172 & $\checkmark$ \\
JBU1015 & 198 & +67 & 369 & +302 & $\checkmark$ \\
JBU1015 & 198 & +115 & 360 & +245 & $\checkmark$ \\
IBE0379 & 470 & +909 & 1644 & +735 & $\checkmark$ \\
DAL1330 & 153 & +238 & 310 & +72 & $\checkmark$ \\
SWA519 & 120 & $-$384 & 241 & $-$143 & $\times$ \\
SWA4348 & 106 & +179 & 232 & +53 & $\checkmark$ \\
JBU1869 & 189 & +73 & 470 & +397 & $\checkmark$ \\
JBU1869 & 196 & $-$326 & 356 & +30 & $\checkmark$ \\
DAL3239 & 212 & +220 & 435 & +215 & $\checkmark$ \\
AAL1498 & 165 & $-$380 & 437 & +57 & $\checkmark$ \\
AAY321 & 67 & $-$36 & 174 & +138 & $\checkmark$ \\
UAL2617 & 177 & +43 & 340 & +297 & $\checkmark$ \\
DAL1523 & 129 & +367 & 302 & $-$65 & $\times$ \\
DAL1072 & 105 & $-$632 & 261 & $-$371 & $\times$ \\
DAL3106 & 67 & $-$42 & 163 & +121 & $\checkmark$ \\
DAL1996 & 154 & +139 & 304 & +165 & $\checkmark$ \\
SCX1419 & 122 & $-$131 & 237 & +106 & $\checkmark$ \\
SWA709 & 95 & $-$833 & 224 & $-$609 & $\times$ \\
SWA4493 & 91 & $-$422 & 193 & $-$229 & $\times$ \\
DAL354 & 296 & $-$111 & 624 & +513 & $\checkmark$ \\
UAL521 & 84 & $-$167 & 202 & +35 & $\checkmark$ \\
FDX457 & 440 & +401 & 1099 & +698 & $\checkmark$ \\
DAL2150 & 197 & $-$2 & 368 & +366 & $\checkmark$ \\
DAL1881 & 218 & $-$92 & 419 & +327 & $\checkmark$ \\
FFT4058 & 163 & $-$810 & 384 & $-$426 & $\times$ \\
SWA1901 & 120 & $-$146 & 279 & +133 & $\checkmark$ \\
JBU1266 & 133 & $-$149 & 349 & +200 & $\checkmark$ \\
JBU1707 & 270 & $-$134 & 521 & +387 & $\checkmark$ \\
SWA4308 & 90 & $-$70 & 219 & +149 & $\checkmark$ \\
\hline
\multicolumn{6}{|l|}{Coverage: 28/36 (77.8\%). $\Delta t = t_{ADS\text{-}B} - t_{FAA}$; Margin $= \gamma\sigma_t - |\Delta t|$.} \\
\multicolumn{6}{|l|}{$^\dagger$Repeated callsigns (e.g., JBU1015) represent distinct flight plan filings on the same date.} \\
\hline
\end{tabular}
\end{table}

Table~\ref{tab:validation} reports the multi-sigma coverage statistics with 95\% confidence intervals. The observed coverages are broadly consistent with theoretical Gaussian predictions at lower sigma levels: at 0.5--1.0$\sigma$, coverage is slightly conservative, while at 3.0$\sigma$ it closely matches theory. At 2.0$\sigma$, the observed coverage (86.1\%) falls below the theoretical value (95.4\%), with the 95\% CI upper bound (93.9\%) not reaching the theoretical target. This suggests the model is slightly under-conservative at the dataset extremes, likely attributable to the small sample size ($N = 36$) and the presence of flights with large ATC-induced deviations that violate the Gaussian assumption. Overall, the model is well-calibrated at the operationally relevant 1-$\sigma$ level used for planning.

\begin{table}[H]
\centering
\caption{Multi-Sigma Validation Statistics ($\gamma = 0.7$, $N = 36$)}
\label{tab:validation}
\begin{tabular}{|c|c|c|c|c|}
\hline
$\sigma$ Level & Theoretical & Observed & 95\% CI & $n$ Covered \\
\hline
0.5$\sigma$ & 38.3\% & 47.2\% & [32.0\%, 63.0\%] & 17/36 \\
1.0$\sigma$ & 68.3\% & 77.8\% & [61.9\%, 88.3\%] & 28/36 \\
1.5$\sigma$ & 86.6\% & 83.3\% & [68.1\%, 92.1\%] & 30/36 \\
2.0$\sigma$ & 95.4\% & 86.1\% & [71.3\%, 93.9\%] & 31/36 \\
3.0$\sigma$ & 99.7\% & 97.2\% & [85.8\%, 99.5\%] & 35/36 \\
\hline
\multicolumn{5}{|l|}{RMSE: 374.5 s, MAE: 275.8 s, Bias: $-$35.5 s} \\
\hline
\end{tabular}
\end{table}

\begin{figure}[H]
    \centering
    \includegraphics[width=\textwidth]{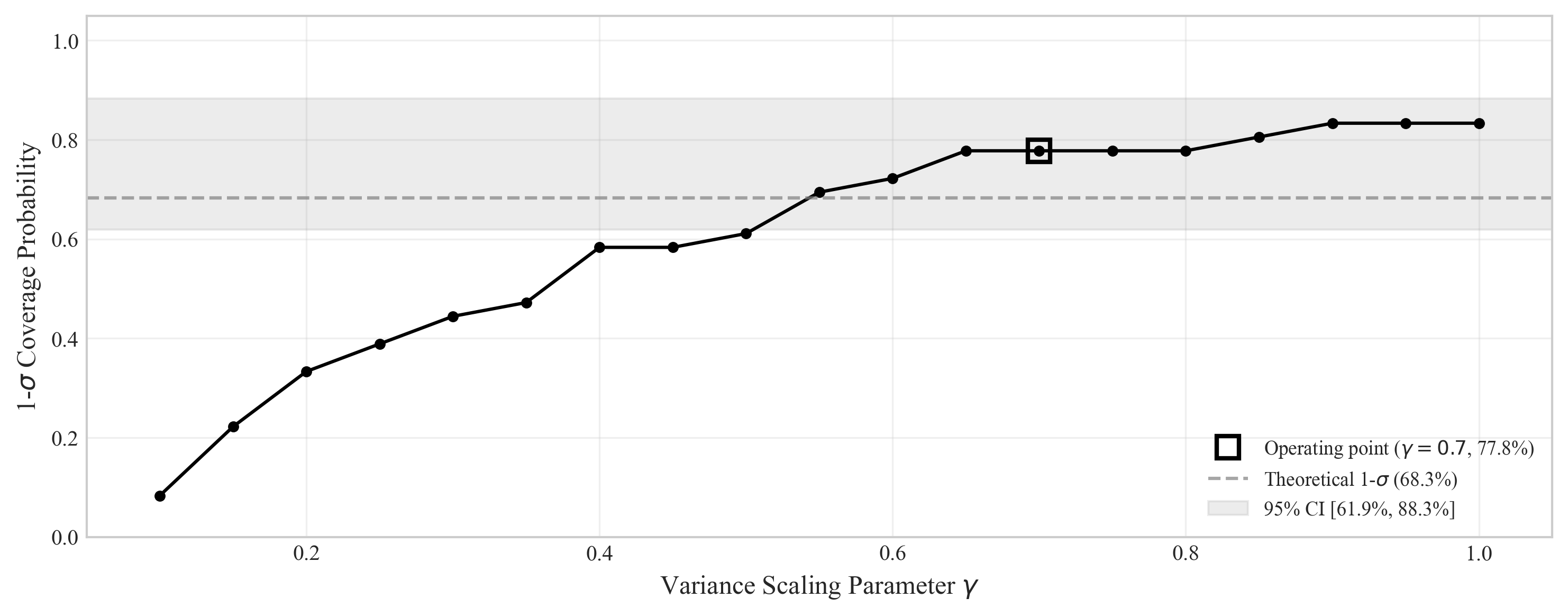}
    \caption{1-$\sigma$ coverage probability as a function of variance scaling parameter $\gamma$. The horizontal reference line indicates the theoretical 68.3\% Gaussian coverage. The operating point $\gamma = 0.7$ is marked.}
    \label{fig:coverage_vs_gamma}
\end{figure}

Figure~\ref{fig:coverage_vs_gamma} shows the relationship between the variance scaling parameter $\gamma$ and 1-$\sigma$ coverage probability. The operating point $\gamma = 0.7$ achieves 77.8\% coverage at the 1-$\sigma$ level on the verification dataset ($N = 36$), consistent with the theoretical 68.3\% Gaussian interval within the 95\% confidence interval [61.9\%, 88.3\%].

\subsection{Comparative Analysis}

The proposed KF-FMS method is compared against four alternative approaches: Monte Carlo simulation without FMS (MC-STD), Monte Carlo with FMS corrections (MC-FMS), LPA constant growth model (LPA-CGM), and LPA with final RTA mode (LPA-RTA). Table~\ref{tab:comparison} summarizes the performance metrics across all tested routes.

\begin{table}[H]
\centering
\caption{Uncertainty Propagation Methods Comparison (Flight Route 3)}
\label{tab:comparison}
\begin{tabular}{|l|c|c|c|c|}
\hline
Method & Time (s) & Final Unc. (m) & RTA Var. (s) & Memory (MB) \\
\hline
MC-STD (1000 samples) & 9.20 & 212,788 & 1478.3 & 17.4 \\
MC-FMS (1000 samples) & 21.69 & 35.7 & 0.20 & 17.3 \\
KF-FMS (proposed) & 0.15 & 90.0 & 50.8 & 4.9 \\
LPA-CGM & 0.10 & 4518.4 & 2259.2 & 4.6 \\
LPA-RTA & 0.11 & 1231.1 & 615.6 & 4.6 \\
\hline
\end{tabular}
\end{table}

The proposed KF-FMS method achieves a final position uncertainty of 90 m at 0.15 seconds computational cost, 60$\times$ faster than MC-STD and 145$\times$ faster than MC-FMS. The MC-FMS method produces tighter bounds (35.7 m) by simulating explicit FMS corrections across 1000 trajectories, but at 145$\times$ the computational cost. The LPA methods are marginally faster but produce substantially more conservative bounds (1231--4518 m) because they model uncertainty growth as linear or piecewise without accounting for progressive FMS correction. MC-STD without FMS corrections produces unrealistically large bounds (212,788 m), confirming the importance of modeling FMS behavior.

\subsection{Controlled Covariance Recovery}

A controlled ablation was conducted to verify that the KF propagation produces correct uncertainty bounds as a property of the filter itself, independent of the real-data $Q_{max}$ calibration. Given a specified process noise intensity $\sigma_a$, we compare the KF-predicted destination position uncertainty $\|\sigma_{pos}\|$ against the empirical RMS estimate-error from $N = 500$ Monte Carlo realizations of the underlying state-space system. In each realization, the ``true'' state evolves under the assumed dynamics with process noise $w_k \sim \mathcal{N}(0, R(\sigma_a))$, and a KF instance receives position observations with measurement noise $\nu_k \sim \mathcal{N}(0, Q(p_k))$. Because $Q_{true}$ is specified rather than learned from data, this ablation tests the filter's covariance propagation math in isolation, avoiding the circular-validation trap of tuning and testing on the same synthetic source.

\begin{figure}[H]
    \centering
    \includegraphics[width=\textwidth]{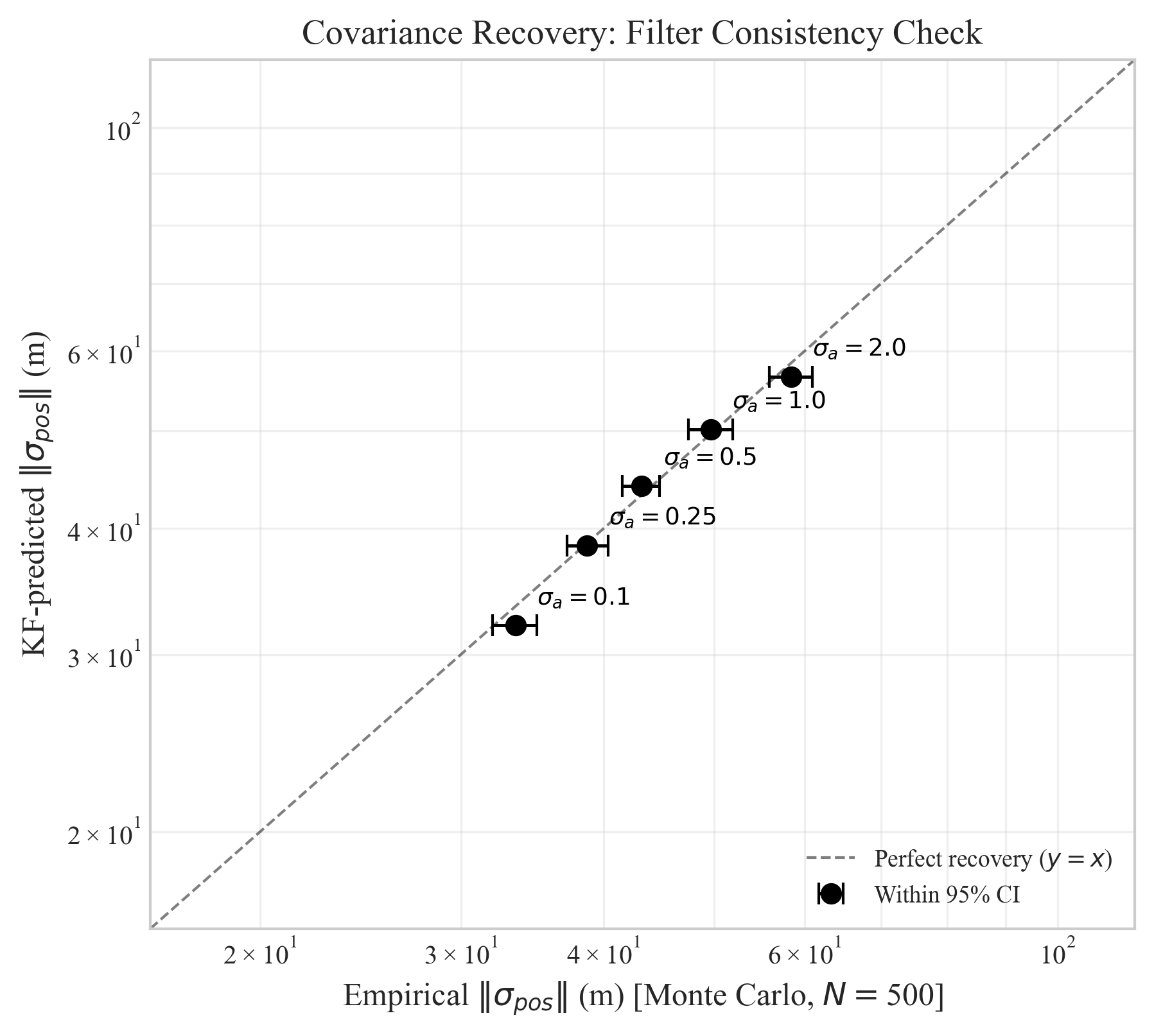}
    \caption{KF-predicted destination position standard deviation $\|\sigma_{pos}\|$ vs.\ empirical RMS estimate-error from $N = 500$ Monte Carlo realizations, at five process-noise intensities. Horizontal error bars: 95\% bootstrap CI on the empirical RMS. Dashed line: perfect recovery ($y = x$).}
    \label{fig:cov_recovery}
\end{figure}

Figure~\ref{fig:cov_recovery} shows the result across five $\sigma_a$ intensities spanning a factor of 20. Table~\ref{tab:cov_recovery} lists the full numerical values. At all five intensities, the predicted $\|\sigma_{pos}\|$ lies within the 95\% CI of the empirical RMS, with recovery ratios $\sigma_{pred}/\sigma_{emp} \in [0.97, 1.02]$. This confirms that the sigmoid-blended KF correctly transports injected process-noise covariance to the predicted state covariance across nearly two orders of magnitude in $\sigma_a$, consistent with classical Kalman filter theory and independent of the empirical $Q_{max}$ calibration addressed in Section~\ref{sec:validation}.

\begin{table}[H]
\centering
\caption{Covariance Recovery Ablation Results ($N = 500$ MC per intensity)}
\label{tab:cov_recovery}
\begin{tabular}{|c|c|c|c|c|}
\hline
$\sigma_a$ & Predicted (m) & Empirical (m) & 95\% CI (m) & Within CI \\
\hline
0.10 & 32.1 & 33.5 & [32.0, 35.0] & Yes \\
0.25 & 38.5 & 38.7 & [37.1, 40.4] & Yes \\
0.50 & 44.1 & 43.2 & [41.5, 44.8] & Yes \\
1.00 & 50.2 & 49.6 & [47.5, 51.9] & Yes \\
2.00 & 56.6 & 58.4 & [55.9, 60.9] & Yes \\
\hline
\end{tabular}
\end{table}

\subsection{Limitations and Applicability}

The model was calibrated using commercial IFR flight data (Boeing 737-series and Airbus A3xx-series aircraft). While these aircraft share the key property of automated FMS tracking of filed flight plans with future UAM vehicles, direct application to UAM operations requires re-calibration of $Q_{max}$ with UAM-specific flight data once such data becomes publicly available. The following differences between commercial IFR and UAM operations are relevant:

\begin{itemize}
    \item UAM vehicles operate at lower altitudes (500--2000 ft AGL) where atmospheric turbulence is typically higher, suggesting larger $\sigma_a^2$ values.
    \item UAM flights are shorter (10--30 minutes) compared to the 1--10 hour commercial flights used for calibration, reducing the cumulative effect of uncertainty growth.
    \item UAM vehicles may employ more responsive FMS systems with faster correction rates, suggesting that $Q_{max}$ may be lower (tighter tracking) than observed in commercial operations.
\end{itemize}

The framework's parametric nature makes it adaptable: $Q_{max}$, $\sigma_a^2$, and $\gamma$ can be independently tuned for any aircraft class given sufficient validation data. The sensitivity analysis in Section~\ref{sec:results} demonstrates that the model responds predictably to parameter changes, supporting its applicability across different operational regimes.

% Comments
% What are predicted confidence levels? What do they mean? Figure 13 Section C. What are coverage fractions? Why do points near the diagonal indicate well-calibrated predictions. These need to be explained more.
\section{Conclusion}
\label{sec:conclusion}

This paper presented a Kalman Filter-based uncertainty propagation method for modeling FMS correction behavior in strategic flight planning. The primary contribution is a sigmoid-blended measurement noise covariance that generalizes the discrete LPA activation threshold into a continuous, parametrically tunable function. This formulation enables the Kalman Filter to serve as a computationally efficient surrogate for FMS trajectory tracking, producing physically informed uncertainty profiles that grow during uncorrected flight and contract as waypoints are approached.

The method was calibrated using real ADS-B data from commercial IFR flights and verified on an independent held-out dataset (of size $N = 36$). At the operationally relevant 1-$\sigma$ level, the model achieved 77.8\% coverage probability, relative to the theoretical 68.3\% Gaussian target. The KF-FMS method is 60--145$\times$ faster than Monte Carlo alternatives while producing bounded, non-conservative uncertainty estimates, a favorable trade-off for real-time strategic planning services where thousands of flight plans must be evaluated per planning cycle.

The framework is parametrically adaptable: $Q_{max}$, $\sigma_a^2$, and $\gamma$ can be independently re-calibrated for different aircraft classes or operational environments. As UAM flight data becomes available, direct calibration for eVTOL platforms will replace the commercial IFR surrogates used here.

\bibliography{bibliography}

\end{document}